\documentclass[letterpaper]{article} 
\usepackage{aaai2027}  
\usepackage[hyphens]{url}  
\usepackage{graphicx} 
\usepackage{natbib}  
\usepackage{caption} 
\usepackage{algorithm}
\usepackage{algorithmic}
\usepackage{amsmath}
\usepackage{multirow}
\usepackage{amssymb}
\usepackage{newfloat}
\usepackage{listings}
\DeclareCaptionStyle{ruled}{labelfont=normalfont,labelsep=colon,strut=off} 
\floatstyle{ruled}
\newfloat{listing}{tb}{lst}{}
\floatname{listing}{Listing}

\usepackage{booktabs}

\title{HCPN-GCN: Scaling Hierarchical Prototype Networks with Cone Geometry for Continual Graph Learning}

\title{HCPN-GCN: Scaling Hierarchical Prototype Networks with Cone Geometry for Continual Graph Learning}
\author {
    Sammuel R. Silva\textsuperscript{\rm 1}\equalcontrib,
    Vander L. S. Freitas\textsuperscript{\rm 1},
    Gladston Moreira\textsuperscript{\rm 1},
    Eduardo J. S. Luz\textsuperscript{\rm 1},
    Rodrigo Silva\textsuperscript{\rm 1}
}
\affiliations {
    \textsuperscript{\rm 1}Universidade Federal de Ouro Preto, Ouro Preto 35402-163, Brasil\\
    sammuel.silva@aluno.ufop.edu.br, vander.freitas@ufop.edu.br, 
    gladston@ufop.edu.br,
    eduluz@ufop.edu.br,
    rodrigo.silva@ufop.edu.br
}

\begin{document}

\maketitle

\begin{abstract}
Continual Graph Learning (CGL) aims to incrementally learn from graph-structured data while preserving knowledge acquired from previous tasks. A major challenge in this setting is catastrophic forgetting, where learning new tasks degrades performance on previously learned ones. Hierarchical Prototype Networks (HPNs) address this problem through a prototype-based memory mechanism that avoids storing historical data, but their reliance on linear feature extractors limits their ability to exploit graph topology, while point-based prototypes often lead to inefficient prototype growth on structurally diverse graphs. In this work, we propose HCPN-GCN, a graph-aware extension of HPN that replaces the original linear feature extractors with Graph Convolutional Networks (GCNs) and introduces cone-based prototypes with a diversity regularization objective. The proposed design produces richer graph-aware representations while compactly modeling the embedding space, reducing prototype proliferation without sacrificing discriminability. Experimental results on six continual graph learning benchmarks demonstrate that HCPN-GCN consistently improves average classification accuracy over the original HPN and representative continual learning baselines while maintaining near-zero forgetting. Furthermore, our analysis shows that the proposed model learns substantially richer class-level prototype hierarchies using approximately $30\times$ fewer atomic prototypes than the original HPN, providing a more compact and effective memory representation for continual graph learning.

\end{abstract}


\section{Introduction}
Graph Neural Networks (GNNs) are the dominant approach for learning from graph-structured data such as social networks, citation networks, and molecular systems \cite{liu2020overcomingcatastrophicforgettinggraph,JIANG2026132566,Yuan_2026,shen2023graphrelationawarecontinual,Bhagat_2011,DBLP:journals/corr/KipfW16}, yet most GNNs assume the entire graph is available during training. In practice, graphs evolve over time as new nodes, edges, or classes are continuously introduced, causing the underlying data distribution to change.

A straightforward solution is to retrain the model whenever new data become available. In practice, however, repeated retraining is often infeasible due to computational cost, privacy restrictions that prevent storing historical data, and applications requiring online adaptation. These challenges motivate \emph{Continual Graph Learning} (CGL), in which a model must incrementally learn new graph data while preserving knowledge acquired from previous tasks. The primary challenge in this setting is catastrophic forgetting, where optimizing for new tasks degrades performance on previously learned ones \cite{Kirkpatrick_2017}.

\begin{figure}[t]
    \centering
    \includegraphics[width=0.85\linewidth]{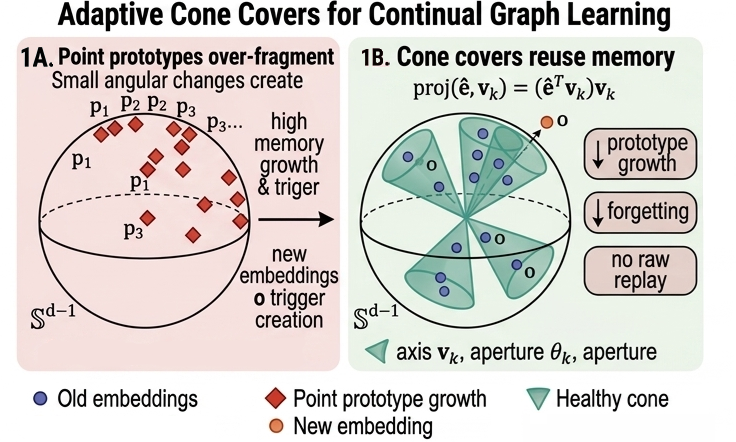}
    \caption{Point prototypes over-fragment under small angular changes (1A), while cone covers reuse existing memory via projection, reducing prototype growth, forgetting, and the need for raw replay (1B).}
    \label{fig:motivation}
\end{figure}

Existing continual graph learning methods generally follow three strategies. Replay-based methods alleviate forgetting by revisiting stored samples from previous tasks, but require additional memory and may violate privacy constraints. Regularization-based approaches preserve important parameters through optimization penalties, although their effectiveness typically decreases as the number of tasks grows. Parameter-isolation methods avoid interference by assigning dedicated model capacity to different tasks, providing stronger forgetting resistance at the expense of increased architectural complexity \cite{CaT,10.5555/3295222.3295393,zhou2021overcomingcatastrophicforgettinggraph,liu2020overcomingcatastrophicforgettinggraph,Zhang23TPAMI,Zhang_2023}.

Among parameter-isolation approaches, the Hierarchical Prototype Network (HPN) \cite{Zhang23TPAMI} represents knowledge using a hierarchy of atomic, node, and class prototypes instead of storing historical data. During continual learning, only prototypes activated by incoming samples are updated, while inactive prototypes remain unchanged, effectively mitigating catastrophic forgetting. Although HPN demonstrates strong continual learning performance, it presents two important limitations. First, node representations are generated using linear Atomic Feature Extractors (AFEs), which cannot fully exploit the relational information available in graph topology. Second, each prototype is represented by a single point in the embedding space, requiring many prototypes to model structurally diverse graphs and leading to rapid growth of the prototype vocabulary (Figure~\ref{fig:motivation}).

To address these limitations, we propose HCPN-GCN, a graph-aware extension of HPN that improves both representation learning and prototype efficiency. We replace the original linear AFEs with Graph Convolutional Networks (GCNs), enabling iterative neighborhood aggregation through message passing and producing richer graph-aware embeddings. Furthermore, we replace point prototypes with cone-based prototypes, allowing each prototype to represent a region of the embedding space rather than a single point. Combined with a diversity regularization objective and a controlled prototype expansion strategy, this formulation substantially reduces prototype proliferation while preserving discriminative power.

The main contributions of this work are summarized as follows:

\begin{itemize}
\item We propose HCPN-GCN, a graph-aware extension of Hierarchical Prototype Networks that replaces linear Atomic Feature Extractors with Graph Convolutional Networks for continual node classification.

\item We introduce cone-based prototype representations together with a diversity regularization mechanism, significantly reducing prototype growth while preserving discriminative representations.

\item We experimentally demonstrate that HCPN-GCN improves continual learning performance across multiple benchmark datasets and provides a substantially more compact prototype hierarchy, requiring approximately $30\times$ fewer atomic prototypes than the original HPN.
\end{itemize}

\section{Related works}
\subsection{Continual Graph Learning}
Continual Graph Learning is a field whose primary objective is to mitigate the effects of catastrophic forgetting \cite{cho2026forgetforgettingcontinuallearning}, a phenomenon in which a model’s performance on previously learned tasks deteriorates as new tasks are introduced. Existing approaches to this problem are commonly grouped into three main categories: replay-based methods, regularization-based methods, and parameter-isolation methods.

Replay-based methods mitigate catastrophic forgetting by storing previously observed samples, or generating synthetic ones, and replaying them alongside new data during training \cite{Yuan_2026, zhang2024}. Gradient Episodic Memory (GEM) \cite{10.5555/3295222.3295393} stores representative samples and constrains gradient updates to preserve prior performance, CAT \cite{CaT} generates condensed synthetic graphs for efficient replay, and ER-GNN \cite{zhou2021overcomingcatastrophicforgettinggraph} reduces memory requirements by storing only node attributes.

Regularization-based methods address catastrophic forgetting by directly constraining changes to previously learned parameters during training on new tasks. Since forgetting mainly occurs when important weights are overwritten, these methods introduce penalty terms that preserve knowledge acquired from earlier tasks. Elastic Weight Consolidation (EWC) \cite{Kirkpatrick_2017} estimates parameter importance using the Fisher Information Matrix and penalizes significant updates to critical weights. Topology-Aware Weight Preserving (TWP) \cite{liu2020overcomingcatastrophicforgettinggraph} extends this idea to graph data by explicitly preserving topological knowledge learned from previous tasks. Memory Aware Synapses (MAS) \cite{10.1007/978-3-030-01219-9_9} measures parameter importance based on the sensitivity of the model outputs to weight changes.

Parameter-isolation methods allocate dedicated parameters to each new task, preventing cross-task interference. PI-GNN \cite{Zhang_2023} expands task-specific subnetworks for graph data, while ContinualGNN \cite{continualGNN} combines isolation with regularization. The Hierarchical Prototype Network (HPN) \cite{Zhang23TPAMI} alleviates the model-growth cost by dynamically expanding a shared pool of feature extractors and selectively activating only the components relevant to each task.

\section{Hierarchical Cone Prototype Networks}
In this section, first we present the problem formulation, and then the HCPN-GCN architecture. HCPN-GCN extends the Hierarchical Prototype Network with two architectural contributions: a Graph Convolutional Network (GCN) backbone that replaces the linear Atomic Feature Extractors of the original HPN, and a cone prototype representation that replaces point-vector prototypes at the atomic and object levels of the hierarchy.

\subsection{Problem Formulation}
In this work, we address the Task-Incremental Learning (Task-IL) problem in the context of Graph Continual Learning. Given a sequence of $p$ tasks ${\mathcal{T}_i \mid i = 1, \dots, p}$, each task $\mathcal{T}_i$ introduces a subgraph $\mathcal{G}_i = (\mathcal{V}_i, \mathcal{E}_i, \mathbf{X}_i)$, where $\mathcal{V}_i$ is a set of nodes belonging to a new set of categories $\mathcal{C}_i$, $\mathcal{E}_i$ denotes the edges among nodes in $\mathcal{V}_i$, and $\mathbf{X}_i \in \mathbb{R}^{|\mathcal{V}_i| \times d_v}$ is the matrix of node feature vectors, with $d_v$ denoting the raw input feature dimension. The category sets are disjoint across tasks, i.e., $\mathcal{C}_i \cap \mathcal{C}_j = \emptyset$ for $i \neq j$.

The model is trained sequentially on $\mathcal{T}_1, \dots, \mathcal{T}_p$, with access only to the current task's data at each step, prior task data is not retained. At test time, the task identity $i$ is provided alongside the input, and the model must classify nodes among the categories $\mathcal{C}_i$ of the corresponding task. A well-performing model must therefore satisfy two competing objectives: \textit{plasticity}, the ability to acquire knowledge from each new task, and \textit{stability}, the ability to retain performance on all previously seen task.

\subsection{GCN-based Atomic Feature Extractors}

Let $N = |\mathcal{V}_i|$ denote the number of nodes in the current task's subgraph. In the original HPN \cite{Zhang23TPAMI}, each Atomic Feature Extractor is a learnable linear transformation $\mathbf{A}_i \in \mathbb{R}^{d_v \times d_a}$ applied directly to node feature vectors, without any aggregation over the graph topology, where $d_a$ is the atomic embedding dimension. We replace this with a \textbf{multi-head GCN} that performs one round of neighborhood aggregation:

\begin{equation}
\mathbf{H} = \sigma\left(\hat{\mathbf{A}}, \mathbf{X}, \mathbf{W}\right),
\end{equation}

\noindent where $\hat{\mathbf{A}}$ is the symmetrically normalized adjacency matrix with self-loops, $\mathbf{X} \in \mathbb{R}^{N \times d_v}$ is the node feature matrix, $M$ is the total number of heads in the AFE pool, $\mathbf{W} \in \mathbb{R}^{d_v \times (d_a \cdot M)}$ is a shared weight matrix, and $\sigma$ is a nonlinear activation. The output is reshaped to $\mathbf{H} \in \mathbb{R}^{N \times M \times d_a}$, where each of the $M$ heads constitutes one AFE, producing an embedding that encodes both node attributes and local graph structure through message passing.

The weight matrix $\mathbf{W}$ is a trainable parameter updated during backpropagation. Following the HPN selective-activation principle, only the heads selected as relevant for a given node receive gradient updates; the remaining columns of $\mathbf{W}$ are masked, preventing cross-task interference at the feature extractor level.

\subsection{Cone Prototype Representation}
\label{sec:cone_repr}

\begin{figure}[t]
    \centering
    \includegraphics[width=0.85\linewidth]{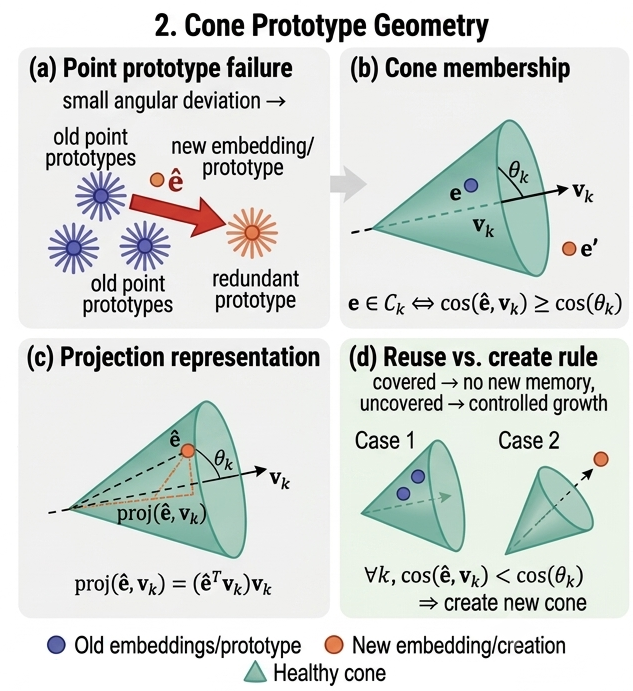}
    \caption{(a) Point prototypes fail under small angular deviation, creating redundant prototypes. (b) An embedding belongs to cone $k$ iff $\cos(\hat{\mathbf{e}}, \mathbf{v}_k) \geq \cos(\theta_k)$. (c) Embeddings are represented via projection onto the cone axis. (d) Covered embeddings reuse existing cones; uncovered embeddings trigger controlled creation of a new cone.}
    \label{fig:cone_geometry}
\end{figure}

We replace each point-vector prototype with a cone, a region of the unit hypersphere $\mathbb{S}^{d-1}$ defined by:
\begin{itemize}
\item An axis $\mathbf{v}_k \in \mathbb{S}^{d-1}$, $\|\mathbf{v}_k\| = 1$, the central direction of the cone.
\item A half-aperture $\theta_k \in \mathbb{R}{>0}$, the angular radius.
\end{itemize}

Cones are created dynamically during training whenever an embedding cannot be assigned to an existing prototype. The axis $\mathbf{v}_k$ and half-aperture $\theta_k$ are then optimized jointly with the network parameters, allowing the prototype vocabulary to grow adaptively as new tasks are encountered.
 
Figure~\ref{fig:cone_geometry} illustrates this geometry and contrasts it with the point-prototype failure mode.

An embedding $\mathbf{e} \in \mathbb{R}^d$ belongs to cone $k$ if and only if:
\begin{equation}
\frac{\mathbf{e}}{\|\mathbf{e}\|} \cdot \mathbf{v}_k \geq \cos(\theta_k).
\end{equation}

\paragraph{Projection-based association.} Rather than passing the axis $\mathbf{v}_k$ directly as the prototype representation (which loses within-cone variation), we associate each embedding with its \textbf{projection} onto the assigned cone axis:

\begin{equation}
\text{proj}(\mathbf{e}, \mathbf{v}_k) = (\hat{\mathbf{e}} \cdot \mathbf{v}_k)\,\mathbf{v}_k, \quad \hat{\mathbf{e}} = \frac{\mathbf{e}}{|\mathbf{e}|}.
\end{equation}

The scalar $\hat{\mathbf{e}} \cdot \mathbf{v}_k \in [\cos\theta_k, 1]$ encodes the alignment strength, nodes near the cone center produce projections close to $\mathbf{v}_k$, while nodes near the boundary produce shorter projections. This preserves within-cone discriminability while maintaining the same output shape as the original point prototype association.

\paragraph{Membership modes.} We support two assignment strategies. In hard mode, each embedding is assigned to its single closest cone ($\arg\max_k \hat{\mathbf{e}} \cdot \mathbf{v}_k$). In soft mode, membership weights $w_k = \text{ReLU}(\hat{\mathbf{e}} \cdot \mathbf{v}_k - \cos(\theta_k))$ are normalized and used to compute a weighted axis $\bar{\mathbf{v}} = \text{normalize}(\sum_k w_k \mathbf{v}_k)$, producing a smoother projection.

\subsection{Hierarchical Cone Prototypes}

Cone prototypes are applied at the Atomic-level and Object-level  of the HPN hierarchy.

\paragraph{Atomic-level (A-cones).} The GCN head embeddings are matched against existing atomic cones. Let $K$ denote the current number of active atomic cones. A new A-cone is created for embedding $\mathbf{e}$ only if it falls outside all existing cones:
\begin{equation}
\forall k \in \{1, \dots, K\}: \quad \hat{\mathbf{e}} \cdot \mathbf{v}_k < \cos(\theta_k).
\end{equation}

This replaces the original cosine threshold condition and ensures creation is triggered only when no existing cone already covers the region. Additionally, a usage gate suppresses creation when the nearest cone is already heavily assigned (i.e., its usage count exceeds a threshold $\tau$), preventing proliferation of redundant cones in dense regions of embedding space. This creation rule is illustrated in Figure~\ref{fig:cone_geometry}(d): a covered embedding reuses an existing cone, while an uncovered one triggers controlled creation of a new cone.

The atomic prototype representation of node $u$ is obtained by projecting its selected AFE head embeddings (indexed by $j$) onto their assigned A-cone axes: $\mathbf{a}(u) = \{\operatorname{proj}(\mathbf{e}_j, \mathbf{v}_{k_j})\}$.

\paragraph{Object-level (O-cones).} The atomic cone projections $\mathbf{a}(u)$ are aggregated via a learned attention mechanism: attention weights are computed over the full atomic projection vector, then used to produce a weighted sum across all atomic projections, yielding a single node-level embedding $\mathbf{o}(u) \in \mathbb{R}^d$, which is $\ell_2$-normalized. This embedding is then matched against existing O-cones using the same cone membership condition, and the same creation and usage-gate logic applies. The object representation is the projection of $\mathbf{o}(u)$ onto its matched O-cone axis.

\paragraph{Class-level.} The object cone projection is $\ell_2$-normalized and passed through a learned linear mapping $\text{FC}_{O \to C}$ to produce a class-level embedding $\mathbf{c}(u)$. Unlike the atomic and object levels, class prototypes are \emph{point vectors}, not cones. This choice is deliberate: the $\text{FC}_{O \to C}$ projection continues to evolve via backpropagation throughout training, making end-of-epoch aperture estimation unreliable (the assigned embedding distribution shifts every epoch as the mapping changes). The two-gate mechanism (Section~\ref{sec:cls_gate}) already constrains class proliferation, so cone geometry at this level would add complexity without a stable aperture update rule.

\subsection{Aperture Adaptation}

Cone apertures are not learned via backpropagation to avoid the circular dependency between assignment and aperture. Instead, they are updated end-of-epoch from the distribution of embeddings assigned to each cone during training. For each active cone $k$, the aperture is set to just cover its outermost assigned embedding with a small margin $\delta$:

\begin{equation}
\cos(\theta_k^{\text{new}}) = \min_{i \in \mathcal{S}_k} \hat{\mathbf{e}}_i \cdot \mathbf{v}_k - \delta,
\end{equation}

\noindent where $\mathcal{S}_k$ is the set of embeddings assigned to cone $k$ during the epoch and $\delta = 0.05$ is a fixed margin. The resulting aperture is clamped to $[\theta_{\min}, \theta_{\max}] = [0.01, \pi/3]$ to prevent degenerate cases. Cones with no assignments during an epoch are left unchanged. Accumulators are reset at the start of each epoch. Cone axes $\mathbf{v}_k$ are updated via gradient descent on the prototype estimation loss; gradients to axes belonging to inactive cones are masked, preventing cross-task interference.

\subsection{Diversity Loss}

To prevent cones from merging or collapsing to points, we apply a two-sided diversity loss at both the atom and object levels:

\begin{equation}
\resizebox{\linewidth}{!}{$
\mathcal{L}_{\text{div}} = w_{\text{ov}} \underbrace{\sum_{i \neq j} \text{ReLU}\left(\mathbf{v}_i \cdot \mathbf{v}_j - \cos(\theta_i + \theta_j)\right)}_{\mathcal{L}_{\text{ov}}} + w_{\text{col}} \underbrace{\sum_k -\log(\theta_k)}_{\mathcal{L}_{\text{col}}}$}
\end{equation}

\noindent where $\mathbf{v}_i, \mathbf{v}_j \in \mathbb{S}^{d-1}$ are unit-norm cone axes (as defined in Section~\ref{sec:cone_repr}), so $\mathbf{v}_i \cdot \mathbf{v}_j = \cos(\angle(\mathbf{v}_i, \mathbf{v}_j))$ exactly. The sum over pairs is $O(K^2)$ in the number of active cones. In practice $K$ remains small (under 200 in all experiments) so runtime overhead is negligible. The first term, $\mathcal{L}_{\text{ov}}$, penalizes pairs of cones whose axes are so close that the cones geometrically intersect (overlap) on the hypersphere. Formally, two cones $i$ and $j$ intersect when the angle between their axes is smaller than the sum of their apertures, which corresponds to $\mathbf{v}_i \cdot \mathbf{v}_j > \cos(\theta_i + \theta_j)$. A ReLU ensures the penalty is zero for non-overlapping pairs and grows linearly with the degree of intersection, pushing cone axes apart until each cone covers a distinct region of the embedding space. The second term, $\mathcal{L}_{\text{col}}$, addresses the opposite failure mode: a cone that shrinks its aperture to zero becomes equivalent to a point prototype (collapse), recovering precisely the degenerate behavior we set out to avoid. The penalty $-\log(\theta_k)$ diverges as $\theta_k \to 0$, providing a hard pressure against collapse without imposing a fixed minimum aperture as a hyperparameter. These two terms are complementary by design: without the overlap penalty, cones could evade intersection by collapsing to points; without the collapse penalty, cones could grow arbitrarily wide and merge into the same region. Together, weighted by $w_{\text{ov}}$ and $w_{\text{col}}$, they maintain a set of cones that are both well-separated and wide enough to provide meaningful coverage of the embedding space.

Class prototypes retain the original cosine threshold penalty from HPN: $\mathcal{L}_{\text{cls-div}} = \sum{i \neq j} \text{ReLU}(\mathbf{c}_i \cdot \mathbf{c}_j - t_{\text{div}})$.

\subsection{Class Creation Gating}
\label{sec:cls_gate}

Class-level prototypes are particularly prone to explosion because the $\text{FC}_{O \to C}$ mapping continues to evolve during training, causing previously stable class embeddings to fall outside existing C-cones and trigger spurious creation. We address this with a two-gate system:

\textbf{Gate 1 — Loss stabilization.} New class prototypes are only created when the standard deviation of the class-level loss over the last $W$ epochs is below a threshold $\epsilon$:
\begin{equation}
g_{\text{stable}} = \mathbf{1}\left[\text{std}(\mathcal{L}_{\text{cls}}^{(t-W:t)}) < \epsilon\right].
\end{equation}

\textbf{Gate 2 — Per-task budget.} A hard cap $B$ limits the number of new class prototypes created within each task. The budget counter resets at each task boundary:
\begin{equation}
g_{\text{budget}} = \mathbf{1}\left[\Delta K_{\text{cls}}^{(\text{task})} < B\right].
\end{equation}

Class creation is enabled only when both gates are open: $g_{\text{stable}} \wedge g_{\text{budget}}$.

\paragraph{Cone lifecycle.} The current implementation supports cone \emph{creation} (described above) but not explicit \emph{merging} or \emph{pruning}. Drift-induced proliferation is controlled passively: the usage gate blocks creation in dense regions, and the diversity loss keeps axes well-separated. If two cones converge so that $\angle(\mathbf{v}_i, \mathbf{v}_j) < \theta_i + \theta_j$, the overlap penalty pushes them apart rather than merging them. In principle, cones with zero usage over an entire task could be pruned without loss of discriminability; we leave this as future work.

\section{Experiments}
In this section, we aim to answer the following research questions: (1) Does replacing linear AFEs with GCN-based AFEs improve continual node classification? (2) What adaptation is required to make GCN-AFEs compatible with continual learning without catastrophic forgetting? (3) How does the hierarchical prototype vocabulary change when the embedding space is graph-structured rather than feature-only?

\subsection{Datasets}
To evaluate the proposed HCPN-GCN model, we conduct experiments on six publicly available graph datasets. Four of these datasets (Cora, Citeseer, Actor, and OGB-Arxiv) were used in the original HPN study~\cite{Zhang23TPAMI}, enabling a direct comparison. In addition, we include CoraFull and arvix-CL ~\cite{zhang2022cglb}, that are widely adopted benchmarks in continual graph learning, to further assess the scalability of the proposed approach on a graph with a greater number of classes and tasks. Table~\ref{tab:datasets} summarizes the main characteristics of the datasets used in our experiments.

\begin{table}[t]
\centering
\caption{Statistics of the benchmark datasets used in the experiments.}
\label{tab:datasets}
\resizebox{\columnwidth}{!}{%
\begin{tabular}{lrrrrrr}
\toprule
\textbf{Statistic} & \textbf{Cora} & \textbf{Citeseer} & \textbf{Actor} & \textbf{OGB-Arxiv} & \textbf{Arxiv-CL} & \textbf{CoraFull} \\
\midrule
Nodes            & 2,708   & 3,327   & 7,600   & 169,343   & 169,343 & 19,793 \\
Edges            & 5,429   & 4,732   & 33,544  & 1,166,243 & 1,166,243 & 130,622 \\
Classes          & 7       & 6       & 4       & 40        & 40 & 70 \\
Tasks            & 3       & 3       & 2       & 20        & 5 & 14 \\
Classes / Task   & 2--3    & 2       & 2       & 2         & 8 & 5 \\
\bottomrule
\end{tabular}%
}
\end{table}

\paragraph{Arxiv-CL vs.\ OGB-Arxiv.}
OGB-Arxiv~\cite{hu2021opengraphbenchmarkdatasets} splits the full Open Graph Benchmark graph (${\sim}$170K nodes, 40 classes) into 20 sequential 2-class tasks.
Arxiv-CL~\cite{zhang2022cglb} uses the same paper graph but applies the CGLB coarser partition: 8 tasks of 5 classes each with fixed benchmark splits.
The two settings differ in task granularity and evaluation protocol and are reported separately.

\subsection{Comparison with Continual Learning Methods}

To evaluate the effectiveness of the proposed HCPN-GCN, we compare its performance against three representative regularization-based continual learning methods: Learning without Forgetting (LwF)~\cite{8107520}, Elastic Weight Consolidation (EWC)~\cite{Kirkpatrick_2017}, Memory Aware Synapses (MAS)~\cite{10.1007/978-3-030-01219-9_9}, and Topology-aware Weight Preserving (TWP)~\cite{liu2020overcomingcatastrophicforgettinggraph}, a method specifically designed for continual graph representation learning. We also compare our approach with the original Hierarchical Prototype Networks (HPN)~\cite{Zhang23TPAMI}, which serves as the foundation of the proposed model. In addition, we report the performance of Joint Training (Joint), in which the model is trained using data from all tasks simultaneously. Although Joint Training is not a continual learning approach, as it assumes access to all data throughout training, it provides an upper-bound performance reference.

Baseline results for LWF, EWC, MAS, TWP, and Joint Training are taken directly from~\cite{Zhang23TPAMI}, using the same graph splits and experimental protocol. The HPN results in Table~\ref{tab:reg_baselines} also correspond to the original implementation, which employs linear Atomic Feature Extractors (AFEs) together with point prototypes. In contrast, our HPN-GCN replaces the original AFEs with a GCN encoder while retaining the hierarchical prototype learning framework.

\paragraph{Evaluation metrics.} We report Average Accuracy (AM) $= \frac{1}{T}\sum_{i=1}^T a_{T,i}$, where $a_{T,i}$ is the test accuracy of task $i$ after all $T$ tasks are trained. Forgetting Measure (FM) $= \frac{1}{T-1}\sum_{i=1}^{T-1}(a_{T,i} - a_{i,i})$, where $a_{i,i}$ is the accuracy of task $i$ right after it was learned. Negative FM ($a_{T,i} < a_{i,i}$) indicates catastrophic forgetting; FM $\approx 0$ indicates stability; positive FM indicates backward transfer, i.e.\ learning later tasks improved performance on earlier ones.

\paragraph{Implementation details of HCPN-GCN.}
All models are trained with Adam (learning rate $0.01$, weight decay $10^{-4}$), using the same optimizer during both the prototype estimation and classification phases. Unless otherwise stated, all experiments use one attribute and two relational AFE heads per node, diversity weights $w_{\text{ov}}{=}0.01$, $w_{\text{col}}{=}0.001$, $\lambda_{\text{div}}{=}0.01$, an initial aperture $\theta_0{=}0.88$\,rad with a 5-epoch warmup, hard membership mode, and class gate window $W{=}5$. Dataset-specific hyperparameters are summarized in Table~\ref{tab:hparams}. Dataset-specific hyperparameters were selected according to graph characteristics and continual learning requirements.

\begin{table}[t]
\centering
\caption{Dataset-specific hyperparameter settings. Hyperparameters common to all datasets are described in the text.}
\label{tab:hparams}
\resizebox{\columnwidth}{!}{%
\begin{tabular}{lcccccc}
\toprule
& \textbf{Cora} & \textbf{Cite.} & \textbf{Actor} & \textbf{OGB-Ax.} & \textbf{CFull} & \textbf{Ax-CL} \\
\midrule
$d$                      & 16      & 16      & 8       & 16      & 16      & 16      \\
GCN hops                 & 3       & 3       & 1       & 2       & 1       & 2       \\
Neighbor samp.         & (5,\,7) & (5,\,7) & (1)     & (5,\,7) & (5,\,7) & (5,\,7) \\
Rel.\ AFE          & 22      & 22      & 22      & 82      & 22      & 44      \\
Epochs (est./cls.)       & 90\,(35/50) & 90\,(35/50) & 90\,(35/50) & 85\,(35/50) & 90\,(35/50) & 110\,(50/60) \\
LR decay                 & 35,\,35 & 35,\,35 & 35,\,35 & 35,\,35 & 35,\,55 & 50,\,75 \\
$\tau$                   & 500     & 500     & 500     & 50      & 500     & 50      \\
$B$                      & 50      & 50      & 50      & 20      & 50      & 25      \\
\bottomrule
\end{tabular}%
}
\end{table}

\begin{table}[t]
\centering
\caption{Performance comparison between representative continual learning methods, the original HPN, and the proposed HCPN-GCN. AM and FM are reported as mean $\pm$ std (\%).}
\label{tab:reg_baselines}

\resizebox{\columnwidth}{!}{%
\begin{tabular}{l cc cc cc cc cc}
\toprule
\multirow{2}{*}{Method}
& \multicolumn{2}{c}{Cora}
& \multicolumn{2}{c}{Citeseer}
& \multicolumn{2}{c}{Actor}
& \multicolumn{2}{c}{OGB-Arxiv} \\
\cmidrule(lr){2-3}
\cmidrule(lr){4-5}
\cmidrule(lr){6-7}
\cmidrule(lr){8-9}
& AM & FM & AM & FM & AM & FM & AM & FM \\
\midrule
EWC
& 63.1$\pm$1.2 & $-42.7\pm$1.6
& 54.4$\pm$2.2 & $-30.3\pm$0.9
& 44.3$\pm$1.1 & $-7.1\pm$1.4
& 72.1$\pm$2.4 & $-9.1\pm$1.9 \\

MAS
& 84.7$\pm$0.7 & $-5.6\pm$2.0
& 69.1$\pm$1.1 & $-4.8\pm$3.3
& 53.7$\pm$2.6 & $-1.6\pm$0.8
& 70.6$\pm$1.3 & $-16.7\pm$1.6 \\

TWP
& 81.3$\pm$3.2 & $-14.4\pm$1.5
& 69.8$\pm$1.5 & $-8.9\pm$2.6
& 54.0$\pm$1.8 & $-2.1\pm$1.9
& 75.8$\pm$0.5 & $-5.9\pm$0.3 \\

HPN
& 93.54$\pm$1.2 & \textbf{+0.7$\pm$1.1}
& 79.51$\pm$0.4 & -0.5$\pm$0.6
& 57.27$\pm$1.0 & -0.7$\pm$1.2
& 85.19$\pm$0.3 & \textbf{+0.8$\pm$1.2} \\

Joint.
& 93.7$\pm$0.5 & -
& 78.9$\pm$0.4 & -
& 57.0$\pm$0.9 & -
& 77.2$\pm$0.8 & - \\

\midrule
\textbf{HCPN-GCN}
& \textbf{93.84$\pm$0.9} & -0.03$\pm$0.5
& \textbf{81.4$\pm$0.7} & \textbf{-0.48$\pm$0.7}
& \textbf{62.3$\pm$0.7} & \textbf{+0.21$\pm$0.7}
& \textbf{86.21$\pm$1.8} & -0.2$\pm$1.3 \\
\bottomrule
\end{tabular}%
}
\end{table}

Table~\ref{tab:reg_baselines} compares HCPN-GCN with representative continual learning baselines and the original HPN. Overall, HCPN-GCN consistently outperforms the regularization-based baselines (EWC, MAS, and TWP) across all evaluated datasets, demonstrating the effectiveness of the proposed HCPN-GCN architecture compared with representative regularization-based continual learning methods.

Compared with HPN, HCPN-GCN achieves comparable performance on Cora (93.84\% vs. 93.54\%) and higher average accuracy on Citeseer (+1.9 \%), Actor (+5.1 \%), and OGB-Arxiv (+1.1 \%). These results indicate that incorporating the proposed graph-convolutional encoder generally improves the discriminative power of the learned prototypes.

Regarding forgetting, HCPN-GCN exhibits competitive performance while preserving previously acquired knowledge. On Citeseer, the proposed method achieves a slightly better forgetting measure than HPN ($-0.48$ versus $-0.5$), while on Actor it obtains a positive FM (+0.21), indicating that later tasks improve average performance on previously learned tasks. On Cora and OGB-Arxiv, HPN still achieves the highest FM values, although HCPN-GCN maintains forgetting close to zero, indicating that the accuracy improvements are obtained without introducing severe catastrophic forgetting.

On OGB-Arxiv, both HPN (85.19\%) and HCPN-GCN (86.21\%) outperform the Joint Training baseline (77.2\%), which is typically treated as an upper bound. As previously noted, all baselines use the same graph splits as defined by~\cite{Zhang23TPAMI}, so data-protocol differences are not the cause. A plausible explanation is that joint optimization over 40 classes from 20 tasks induces severe class imbalance and a harder multi-objective loss landscape: early-stopping on a held-out validation set drawn uniformly across all classes tends to favour the majority tasks. Prototype-based sequential learning, by contrast, dedicates each phase to a small number of classes, reducing within-phase imbalance.

\begin{table}[t]
\centering
\caption{Edge homophily ratio $h$ and AM gain of HCPN-GCN over HPN ($\Delta$AM),
         sorted by increasing homophily. The \#Tasks column highlights that
         OGB-Arxiv operates at substantially larger scale.}
\label{tab:homophily}
\resizebox{\columnwidth}{!}{%
\begin{tabular}{l c c c c r}
\toprule
Dataset     & $h$  & \#Tasks
            & HPN (AM)                    & HCPN-GCN (AM)
            & $\Delta$AM \\
\midrule
Actor       & 0.22 & 2
            & 57.27\scriptsize{$\pm$1.0}   & 62.3\scriptsize{$\pm$0.7}
            & \textbf{+5.1} \\
OGB-Arxiv   & 0.65 & 20
            & 85.19\scriptsize{$\pm$0.5}   & 86.2\scriptsize{$\pm$1.8}
            & +1.0 \\
Citeseer    & 0.74 & 3
            & 79.51\scriptsize{$\pm$0.4}   & 81.4\scriptsize{$\pm$0.7}
            & +1.9 \\
Cora        & 0.81 & 3
            & 93.54\scriptsize{$\pm$1.2}   & 93.8\scriptsize{$\pm$0.9}
            & +0.3 \\
\bottomrule
\end{tabular}%
}
\end{table}

To better understand the performance improvements observed in Table~\ref{tab:reg_baselines}, we analyze the relationship between dataset characteristics and the gains obtained by HCPN-GCN. Table~\ref{tab:homophily} further suggests a relationship between dataset homophily (tendency of connected nodes to share the same class label \cite{Tao_2025}) and the magnitude of improvement. Among the three small-to-medium benchmarks (Actor, Citeseer, and Cora) the observed gains tend to decrease as the edge homophily ratio increases to the edge homophily ratio $h$: the most heterophilic dataset (Actor, $h=0.22$) benefits most from graph-convolutional encoding (+5.1), while the most homophilic (Cora, $h=0.81$) shows a marginal and statistically ambiguous gain (+0.3). This pattern is consistent with the role of GCN-AFEs as neighbourhood encoders: on heterophilic graphs, a node's local structure can provide additional relational information beyond raw node features, and GCN aggregation makes this information available to the prototype matching layer. On highly homophilic graphs, neighbours tend to share the same class, so node features alone are already sufficient for discriminative prototype estimation, and the additional aggregation step provides diminishing returns.

OGB-Arxiv ($h=0.65$, 20 tasks) is a partial exception, yielding only +1.0 pts despite moderate heterophily. With 20 sequential tasks and 40 classes, inter-task prototype interference likely becomes the dominant source of variance, attenuating the encoding benefit. The higher standard deviation on OGB-Arxiv ($\pm$1.8 vs.\ HPN's $\pm$0.5) is consistent with this: larger prototype vocabularies introduce run-to-run variability in the continual learning regime.

\subsection{Component Ablation and Prototype Geometry Analysis}

We conduct a $2\times2$ ablation study on Cora to evaluate the contributions of the GCN encoder and the diversity loss while keeping the proposed cone prototype representation fixed. To isolate the impact of the prototype geometry, we additionally evaluate two variants (A5 and A6) that replace cone prototypes with the original point-prototype formulation while retaining the GCN encoder. Table~\ref{tab:ablation} reports the results over 10 independent runs.

\begin{table}[t]
\centering
\caption{Ablation study on Cora (10 runs). Variants A1--A4 evaluate the GCN encoder and diversity loss with cone prototypes, whereas A5--A6 evaluate the original point-prototype representation. Besides AM and FM, we report the average number of atomic (\#ato), object (\#obj), and class (\#cls) prototypes learned by each variant.}
\label{tab:ablation}
\resizebox{\columnwidth}{!}{%
\begin{tabular}{lcccccccc}
\toprule
Variant & Prot. Type & GCN & Div. & AM & FM & \#ato & \#obj & \#cls \\
\midrule
A1 & Cone & Yes & Yes
& \textbf{93.84$\pm$0.90}
& $-0.03\pm0.50$
& 58.3$\pm$5.4
& 129.1$\pm$17.0
& \textbf{24.4$\pm$12.0} \\

A2 & Cone & No & Yes
& 91.52$\pm$1.01
& +0.28$\pm$0.69
& 62.3$\pm$7.0
& 101.1$\pm$22.2
& 3.7$\pm$4.9 \\

A3 & Cone & Yes & No
& 92.10$\pm$0.96
& +0.07$\pm$0.46
& 65.1$\pm$12.8
& 113.8$\pm$17.6
& 5.2$\pm$6.9 \\

A4 & Cone & No & No
& 92.39$\pm$0.68
& +0.52$\pm$0.48
& 60.3$\pm$8.3
& 89.0$\pm$12.1
& 2.3$\pm$2.4 \\

A5 & Point & Yes & Yes
& 92.04$\pm$0.61
& +0.42$\pm$0.93
& 1675.1$\pm$111.3
& 2679.4$\pm$72.1
& 386.3$\pm$86.2 \\

A6 & Point & Yes & No
& 91.23$\pm$0.55
& +0.332$\pm$0.45
& 1743.8$\pm$61.1
& 2719.6$\pm$42.3
& 422.9$\pm$21.1 \\
\bottomrule
\end{tabular}%
}
\end{table}

Among the cone-based variants (A1--A4), the full model (A1) achieves the highest average accuracy (93.84\%), indicating that the GCN encoder and the diversity loss are complementary. Neither component alone improves upon the baseline configuration (A4), suggesting that the diversity loss is most effective when combined with graph-aware representations.

The point-prototype variants (A5 and A6) further highlight the importance of the proposed cone representation. Replacing cones with point prototypes decreases AM by approximately 1.8 percentage points (A1 vs.~A5) while increasing the number of atomic prototypes from $58.3$ to over $1{,}675$. The point-based variants also learn substantially larger object and class prototype vocabularies, demonstrating that the cone formulation yields a substantially more compact hierarchical representation while achieving higher continual learning accuracy.

Across all variants, richer class-level prototype hierarchies are consistently associated with lower forgetting. Removing the GCN encoder limits the number of class prototypes to fewer than five (A2: $3.7$; A4: $2.3$), whereas enabling it increases this number to $5.2$ (A3) and, together with the diversity loss, to $24.4$ (A1). The point-prototype variants learn hundreds of class prototypes (A5: $386.3$; A6: $422.9$) while achieving lower accuracy, suggesting that simply increasing the number of prototypes does not improve continual learning. Instead, the proposed cone representation yields a substantially more compact and discriminative prototype hierarchy.

\subsection{Prototype Vocabulary Efficiency}

A key advantage of HCPN-GCN is that it learns a substantially more compact prototype vocabulary than the original HPN while achieving higher classification performance. On Cora, the full model (A1) converges to $58.3 \pm 5.4$ atomic prototypes over 10 independent runs, compared with $1{,}628 \pm 312$ atoms reported for the original HPN~\cite{Zhang23TPAMI} (Table 2 of that paper, 10 runs; we quote the mean and our re-reading of the reported variance). This corresponds to nearly a 30$\times$ reduction in vocabulary size while simultaneously improving classification accuracy. Moreover, all ablation variants learn fewer than 65 atomic prototypes, suggesting that the substantial reduction in vocabulary size is primarily attributable to the proposed cone-based prototype representation, rather than the result of a particular architectural component.

\begin{figure}[t]
\centering
\includegraphics[width=\columnwidth]{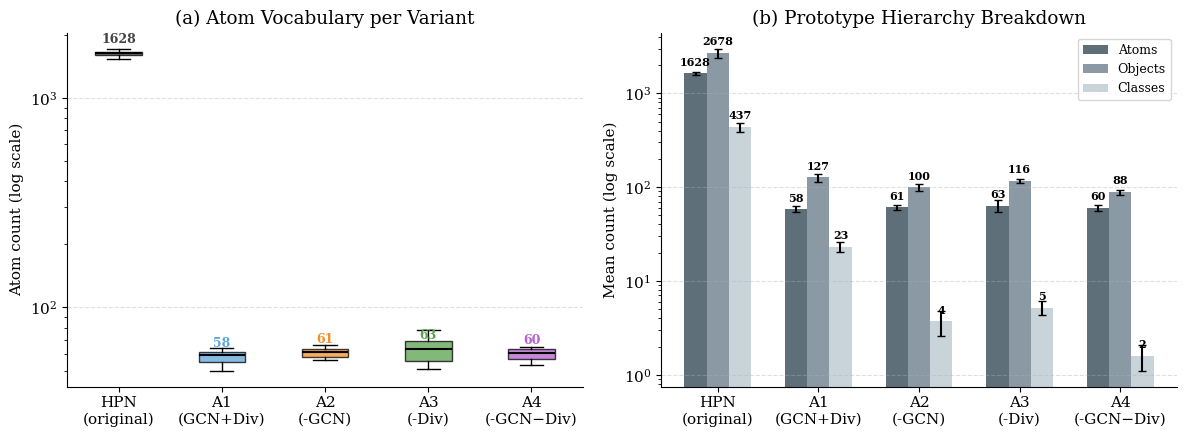}
\caption{Ablation study on Cora. (a) Interaction between the GCN encoder and diversity loss. (b) AM versus FM for the four variants.}
\label{fig:prototypeusage}
\end{figure}

Figure~\ref{fig:prototypeusage} shows that all HCPN-GCN variants converge to fewer than 65 atoms versus 1,628 for HPN. It can be also noted that this compact atomic vocabulary coexists with a richer class-level organisation: the full model (A1) learns $24.4 \pm 12.0$ class prototypes, versus $3.7 \pm 4.9$ (A2) and $2.3 \pm 2.4$ (A4) without the GCN encoder. Graph-aware representations appear to be the main driver of class-level diversity, and that diversity is consistently associated with lower forgetting.

\subsection{Comparison with CGLB-Datasets}

\begin{table}[t]
\centering
\caption{Results on the CGLB benchmark datasets (Arxiv-CL and CoraFull-CL).
Baseline numbers from Zhang et al.~\cite{zhang2022cglb}.
HCPN-GCN results: Arxiv-CL averaged over 7 runs, CoraFull-CL over 3 runs.}
\label{tab:cglb}
\resizebox{\columnwidth}{!}{%
\begin{tabular}{lcccc}
\toprule
& \multicolumn{2}{c}{Arxiv-CL} & \multicolumn{2}{c}{CoraFull-CL} \\
\cmidrule(lr){2-3} \cmidrule(lr){4-5}
Method & AM (\%) & FM (\%) & AM (\%) & FM (\%) \\
\midrule
EWC~\cite{Kirkpatrick_2017}   & 78.8$\pm$2.7 & -5.0$\pm$3.1 & 78.9$\pm$1.7 & -15.5$\pm$2.3 \\
MAS~\cite{10.1007/978-3-030-01219-9_9}       & 88.4$\pm$0.2 & -0.0$\pm$0.0 & 93.0$\pm$0.5 &  -0.6$\pm$0.2 \\
TWP~\cite{liu2020overcomingcatastrophicforgettinggraph}           & 86.0$\pm$0.8 & -2.8$\pm$0.8 & 92.2$\pm$0.5 & -0.9$\pm$0.3 \\
LWF~\cite{8107520}           & 84.2$\pm$0.5 & -3.7$\pm$0.6 & 56.1$\pm$2.0 & -37.5$\pm$1.8 \\
Joint Training               & 90.3$\pm$0.2 & - & 95.2$\pm$0.2 & - \\
\midrule
HCPN-GCN
  & $83.36 \pm 3.2$ & $0.1 \pm 0.2$
  & $94.67 \pm 0.3$ & $-0.00 \pm 0.0$ \\
\bottomrule
\end{tabular}%
}
\end{table}

HCPN-GCN achieves near-zero forgetting on both benchmarks (FM $\approx 0$), indicating that the proposed prototype-based architecture effectively preserves knowledge across sequential tasks on the CGLB benchmark (Arxiv-CL and CoraFull-CL).

On CoraFull-CL, HCPN-GCN achieves an AM of $94.67\%$, outperforming the regularization-based continual learning methods EWC, MAS, TWP, and LwF, while approaching the Joint Training reference ($95.2\%$). At the same time, the model exhibits virtually no forgetting (FM $\approx 0$), demonstrating that the performance gain is obtained without compromising knowledge retention.

On Arxiv-CL, HCPN-GCN obtains an AM of $83.36\%$ with near-zero forgetting (FM $=0.1$). MAS achieves a higher AM ($88.4\%$). Two factors likely contribute to this gap. First, Arxiv-CL's coarser task schedule (8 tasks of 5 classes each, versus OGB-Arxiv's 20 tasks of 2) means each task contains substantially more within-task class overlap, which increases the number of class prototypes created per task and saturates the per-task budget $B$ early in training. Second, the $\text{FC}_{O \to C}$ linear mapping shifts as new tasks are added, causing previously stable class embeddings to drift outside their matched prototypes and requiring creation of redundant class prototypes before the loss-stability gate closes. Both effects reduce classification resolution without contributing to forgetting. Isolating these two factors (budget size and FC drift) with a targeted ablation on Arxiv-CL is a direction for future work.

\section{Conclusion}
We proposed HCPN-GCN, a graph-aware extension of Hierarchical Prototype Networks that replaces linear AFEs with GCN encoders and introduces cone-based prototypes with diversity regularization. Across six continual graph learning benchmark settings, HCPN-GCN consistently improves average accuracy over the original HPN and representative regularization-based baselines while maintaining near-zero forgetting. The ablation study shows that the GCN encoder and the diversity loss are complementary: graph-aware representations promote a richer class-level prototype hierarchy, and that richer hierarchy is consistently associated with lower catastrophic forgetting. The proposed cone-based representation also reduces the atomic prototype vocabulary by approximately $30\times$ relative to HPN, yielding a substantially more compact memory representation.

Future work includes extending the framework to Graph Attention Networks and Graph Transformers, exploring adaptive prototype merging and pruning strategies, and broadening the experimental comparison to include replay-based methods (e.g., CAT and ER-GNN) and dynamic-architecture continual graph learning approaches.

\bigskip

\bibliography{aaai2027}


\end{document}